# Medical Pathologies Prediction : Systematic Review and Proposed Approach

Chaimae Taoussi[*a], Imad Hafidi [†b], and Abdelmoutalib Metrane [‡c]

[a] Laboratory of Process Engineering Computer Science and Mathematics, University Sultan Moulay Slimane, Beni Mellal, Morocco

**Abstract**

The healthcare sector is an important pillar of every community, numerous research studies have been carried out in this context to optimize medical processes and improve care quality and facilitate patient management. In this article we have analyzed and examined different works concerning the exploitation of the most recent technologies such as big data, artificial intelligence, machine learning, and deep learning for the improvement of health care, which enabled us to propose our general approach concentrating on the collection, preprocessing and clustering of medical data to facilitate access, after analysis, to the patients and health professionals to predict the most frequent pathologies with better precision within a notable timeframe.

**keywords:** Healthcare, big data, artificial intelligence, automatic language processing, data mining, predictive models.

## 1 Introduction

Big Data is associated with the massive computational resources required to handle the growing volume and complexity of data from a variety of sources, including structured, semi-structured, or unstructured information. Massive data processing systems necessitate a high level of process automation, which has been aided by the emergence of Artificial Intelligence (AI) in diagnostic decision-making as a result of recent advances in computer technology [1].

In the healthcare sector, the exploitation of multiple sources of big data, including medical records, patient records, and medical examination results, makes advanced methods accessible not only to healthcare professionals but also to their patients [2]. Furthermore, it has been shown that large-scale Electronic Health Records (EHRs) data can totally alter the method of scientific discovery in precision medicine. As a result, EHRs contain invaluable data that should be better utilized to guide biomedical research in a variety of disciplines [3].

Clinical data has lately been subjected to data science techniques, enabling healthcare professionals to create several analytical models. The design and deployment of computer systems that behave as though they understand biomedicine are essential components of knowledge-based biomedical data science [4]. The development of computing in disease detection and diagnosis for biological sciences has occurred throughout the last few decades. Due to the recent exponential growth in applications based on AI-based technologies and the need for doctors to practice with fewer mistakes, mishaps, and misdiagnoses, the medical industry has welcomed and embraced AI [5].

Healthcare professionals need digitalization skills in today's fast-paced environment when using technological equipment or providing technology-enabled services since a lack of competency among healthcare professionals can compromise patient safety and raise the likelihood of errors [6].

Diagnosis of the disease is one of the application areas where data mining techniques aid in the extraction of knowledge from medical databases, and clinical decision support for the prediction of various diseases with good precision. These techniques are very effective in the design of clinical support systems due to their ability to uncover patterns and relationships hidden in medical data [7].

# 2 Methodology:

The methodology of this systematic review adheres to the original format of the PRISMA checklist. A flowchart of the phase is shown in Figure 1.

## 2.1. Identification of studies:

We conducted a detailed literature search, focusing on articles published between 2018 and 2022 to base our analysis on recent studies. A review protocol is designed to minimize researcher bias, this review protocol includes the wording of the research questions, the defined search strategy, and the inclusion and exclusion criteria.

## 2.2. Research questions (RQs):

To clarify the scope of research and be more targeted, RQs have been introduced, which focus on five main points of view such as data collection in medical informatics, data preprocessing in medical informatics, mapping of medical data in medical informatics, classification, and clustering in medical informatics and the prediction of psychological pathologies using artificial intelligence. However, Table 1 represents the RQs considered in this study.

**Table 1** Specific research questions

| | |
|---|---|
| **RQ1** | What are the data collection methods applied in medical informatics? |
| **RQ2** | What are the data preprocessing methods applied in medical informatics? |
| **RQ3** | What are the data mapping techniques applied in medical informatics? |
| **RQ4** | What is the classification and clustering methods applied in medical informatics? |
| **RQ5** | How can artificial intelligence be used to predict medical pathologies? |

## 2.3. Search strategy:

The search strategy focused on the search questions, including consideration of four steps: choosing digital libraries, selecting additional search sources, specifying the time range, and defining search keywords. This study considered the three most popular and extensive online digital libraries: Scopus, PubMed, and Google Scholar. These databases were chosen because of the purpose of their publication, and the purpose is related to our research questions. The search was limited to a period from 2018 to 2022, as we were more interested in recent literature to review the latest knowledge and contributions.

An exhaustive search strategy was followed to find all related articles. The search terms that were used in this search strategy were:

- "Data Collection" **AND** "Health Informatics "
- "Data Preprocessing" **AND** "Health Informatics"
- "Mapping Medical Data" **AND** "Health Informatics "
- "Classification and Clustering" **AND** "Health Informatics "
- "Artificial Intelligence" **AND** " Medical Pathologies Prediction "

## 2.4. Inclusion, exclusion criteria, and data extraction:

A number of articles were subsequently produced as a result of the earlier processes. Thus, we defined inclusion and exclusion criteria to sort out irrelevant papers:

- Studies were included if the paper described data collection or preprocessing methods in health informatics, or presented data mapping or clustering techniques in health informatics. Finally, the paper is eligible if it proves the application of artificial intelligence for the prediction of medical pathologies.

- Studies were excluded if the research questions are irrelevant, the papers are literature reviews, studies for which we could not obtain the full text, or if the papers studied had no attention to answering our research questions related to the collection, pre-processing, mapping, and classification of health data in medical informatics as well as the use of artificial intelligence models for the prediction of medical pathologies.

## 2.5. Results:

Our search identified 734 papers published between 2018 and 2022, when analyzing the significance of the research issues, 614 papers were excluded based on title or abstract (Figure 1), leaving 169 papers eligible. However, based on the exclusion criteria 120 papers were excluded, of which 33 contain irrelevant research questions, 28 have no attention to present mapping medical data in health informatics, 18 pay no attention to presenting data collection in health informatics, 15 duplicate papers, 10 methodological papers presenting a new method, 7 pay no attention to presenting data preprocessing in health informatics, 6 pay no attention to present clustering in health informatics, 3 short communications, corrigendum, 3 validation studies. Thus, 49 articles were ultimately selected for this systematic review as shown in Figure 1. In the sections below, we discuss each of the research questions for data analysis in more depth.

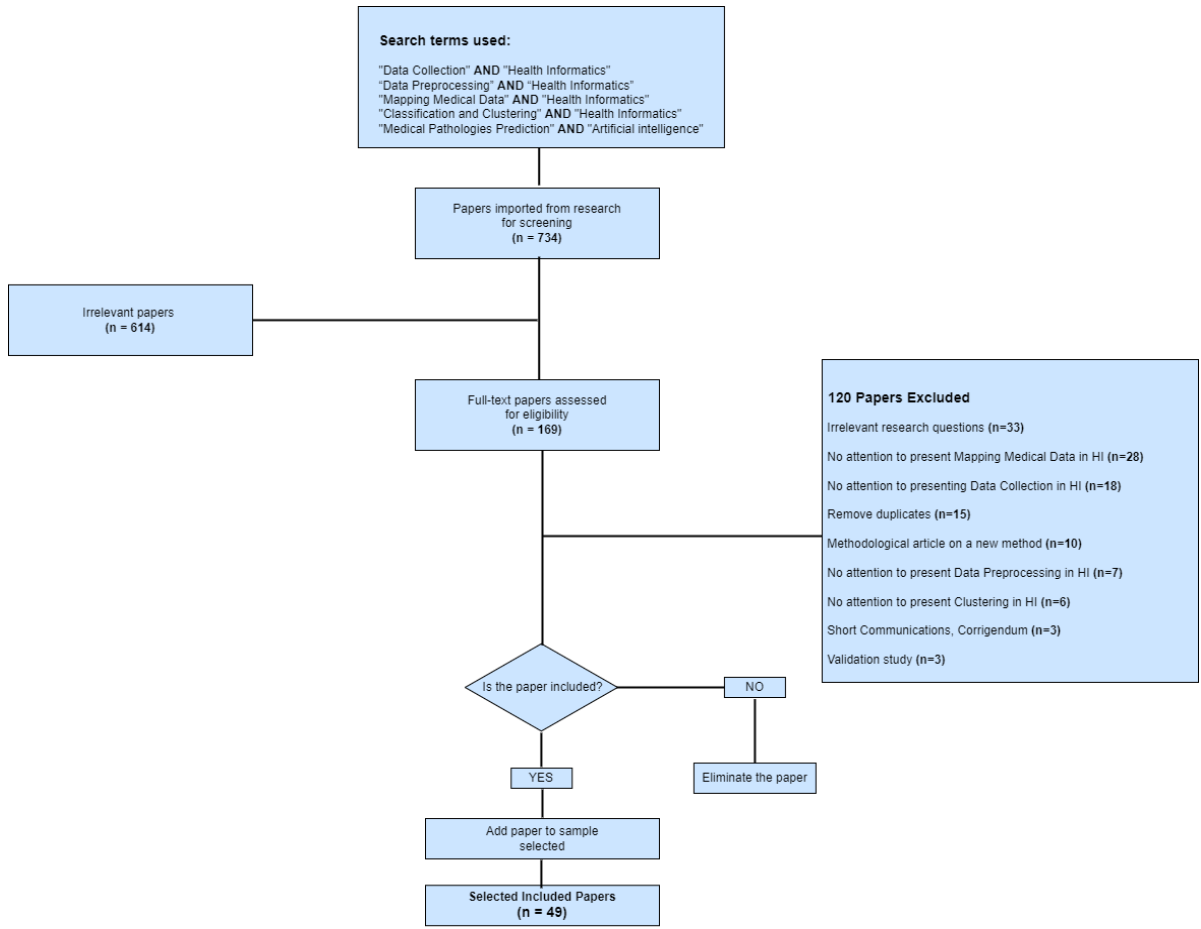

Figure 1: Organigramme PRISMA

## 2.6. Report review:

### 2.6.1. RQ1: What are the data collection methods applied in medical informatics?

Data collection is an essential first step because it significantly improves prevention, diagnosis, treatment, and medical follow-up in the health sector, Seol et al [8] developed an Natural language processing (NLP) solution to detect medical events in electronic medical records for epidemiological purposes by selecting EHRs separated into two file types, Extensible Markup Language (XML) files of textual documents, and XML files containing the metadata of the medical record. This was done by applying a global data model to combine data from various data sources.

The potential of big data in healthcare resides in its capacity to identify patterns and transform massive data sets into insights that can be used by decision-makers and in precision treatment. According to Hariri et al [9], starting with the collecting of individual data items and progressing to the synthesis of heterogeneous data from various sources would reveal whole new techniques to minimize complexity, enhancing illness prediction and prevention.

## 2.6.2. RQ2: What are the data preprocessing methods applied in medical informatics?

Preparing data and reducing complexity requires a necessary step called data pre-processing, of which we can speak about several interventions in many scientific and medical domains. The Natural Language Toolkit (NLTK) is the most effective way to get started in the NLP area. Natural language processing, computational linguistics, and artificial intelligence researchers and students are encouraged by the NLTK set of application packages [10]. Considering the first intervention as a starting point, Jha et al [11] who presented a straightforward method for converting text to emoticons and vice versa using NLTK and WordNet believe that sentiment analysis in the text is a significant field of research in natural language processing. In a similar context, Yao et al [12] developed an algorithm using the NLTK library that allowed it to roughly determine the emotional scores of various sentences and compare them to the scores provided by the person who came up with them.

NLP is becoming more and more important as a result of the quantity of medical information available on a global scale and the growing significance of comprehending and mining big data in the medical industry [13]. A review of the latest relevant contributions to the application of natural language processing techniques to electronic patient records in the free text was presented by ASSALE et al [14].

Artificial intelligence (AI) use in e-health has gained attention in recent years. The use of intelligent interfaces to increase patient engagement and treatment compliance as well as predictive modeling to manage the flow are a few examples of how AI can be used to analyze unstructured data. Carchiolo et al [15] have proposed a system that first extracts the text from the digitized medical prescription before applying the techniques of natural language processing and machine learning.

Data mining is crucial for identifying emerging trends in healthcare. Finding significant insights from massive datasets is one of the most inspiring fields of research. EHR data is made up of high-dimensional, multivariate information, and statistical models are created using data mining approaches. These kinds of data analyses are difficult and susceptible to errors [16]. By incorporating data mining into data frameworks, healthcare organizations may make decisions with less effort while gaining important new medical insights. The goal of predictive data mining in medicine is to create an efficient predictive model, offer accurate predictions, and assist doctors in optimizing their diagnostic and treatment planning processes [17].

The General Architecture for Text Engineering (GATE), which is an architecture, framework, and development environment for linguistic engineering, has gained popularity recently in the context of processing medical data as well, particularly in information extraction from documents written in English. To a certain extent, it also supports other languages, mainly because of the dictionaries that its numerous users build and then share on the platform [18]. Since GATE is one of the well-known open-source tools for text analysis that is accessible in a variety of formats and sizes, in the GATE paradigm, Amjad et al [19] have proposed a brand-new plugin called Unified Approach for Multilingual Sentiment Analysis (UAMSA), which is capable of performing numerous tasks for multilingual sentiment analysis with just one tool. Additionally, UAMSA assists users in accurately classifying the sentiments of people sharing their reviews in multiple languages.

In addition to GATE, PEZOULAS et al [20] developed an automated system for medical data curation and cleaning based on the detection and monitoring of inconsistencies, missing values, anomalies, and similarities, as well as data standards to enable data harmonization. Thus, GOLDBERG et al [21] noted the possibility of using natural language processing and machine learning to anticipate a significant component in the psychotherapy process.

### 2.6.3. RQ3: What are the data mapping techniques applied in medical informatics?

Electronic health records (EHRs), a particular type of health information technology, are being implemented by specialties and clinical settings more often. Topaz et al [22] have developed a new quick clinical text mining system called NimbleMiner, using contemporary NLP methods to enable the mapping of biomedical terminology. Growing volumes of textual data necessitate innovative approaches to data processing and analysis. Al-Hroob et al. [23] proposed a novel method to automatically identify actors and actions in a description based on the requirements of a natural language system. There are discussions of how NLP is crucial in extracting people and actions as well as how ANNs can be used to provide specific identification. Furthermore, WANG et al [24] describe a recently published study on clinical Information Extraction (IE) applications, that automatically extract and encode clinical information from the text to facilitate the secondary use of electronic medical records data.

One of the most effective collaborative efforts in providing resources for biomedical terminology is the Unified Medical Language System (UMLS). In addition to presenting a detailed bibliometric profile of the published UMLS literature and a systematic method for mapping the intellectual development of science, Kim et al [25] also made some suggestions for future possible directions. Still, within the framework of the mapping of medical data, GORRELL et al [26] introduced the new Bio-YODIE system, which consists of two main components: the pipeline that annotates documents that contain a UMLS Concept Unique Identifier (CUI) along with other pertinent UMLS data, and the component in charge of preparing the resources that process the UMLS and the other necessary information resources at runtime. In the same context, we retain the intervention of ABBAS et al [27] who proposed an algorithm that implements the UMLS Terminology Services (UTS) and personalized it to extract concepts for all the expressions and terms used in recitals and determine their semantic and entity types to find an exact categorization of the concepts. This led us to the conclusion that multiple information extraction methodologies, in combination with UMLS, have been used to annotate and extract clinically significant information from medical data sources in several domains of medicine, such as psychology [28] and Cancer [29].

### 2.6.4. RQ4: What is the classification and clustering methods applied in medical informatics?

In the medical field, clustering is used to identify subgroups or groups of patients with a similar profile, to classify heterogeneous patient populations based on their disease diagnostic histories, and to identify phenotypic clusters and their risk factors, Maurits et al [30] conducted a study. The goal of this study was to provide a generalizable framework for the use EHRs collected longitudinally, which will make it easier to conduct research and provide treatment for more homogeneous disease populations. In a different study, Ricciardi et al [31] suggested a machine learning approach that makes use of random forest to identify patients with Parkinson's disease in various stages of the condition based on gait characteristics as opposed to comparing affected and unaffected patients.

Using the Latent Dirichlet Allocation (LDA), an unsupervised probabilistic generative model that falls under the category of thematic models, Wang et al [32] examined the use of unsupervised machine-learning techniques to identify patient subgroups and latent disease clusters in EHR data. They also suggested the Poisson Dirichlet Model (PDM), a novel unsupervised machine learning technique.

In order to enhance the text classification system's performance, Kadhimet al [33] proposed a system for text preprocessing, feature extraction based on the thresholding parameter to classify texts into one or more categories, and they proved that TF-IDF with cosine similarity score performed better in classification. In the same context, Kashina et al [34] examined how each stage of text preparation affects the classification outcome and found that the most precise and straightforward method is logistic regression.

The development of big data categorization approaches is necessary due to the massive generation of medical data in recent years. After realizing that the classification of medical data can be used to visualize hidden patterns and determine the presence of disease from medical data, Jerlin et al [35] presented a successful Multi-Kernel Support Vector Machine (MKSVM) and a Fruit Fly Optimization Algorithm (FFOA) for disease classification. In the same framework, Huang et al [36] proposed a Community-Based Federated Learning (CBFL) algorithm that combined EMR data from various communities and concurrently trained one model per community, making the learning process notably more effective than the Federated Machine Learning (FL) algorithm. Additionally, Desai et al [37] separately investigated genomic datasets and medical imaging data with clinical data points for classification tasks using combinatorial deep learning architectures.

Clinicians would benefit from having access to prediction models for diagnosis without violating patient privacy. Computing encrypted medical data is possible using Fully Homomorphic Encryption (FHE), which also protects data confidentiality. A cryptographic technique for Naive Bayes private classification was developed by Wood et al [38] using completely homomorphic private key encryption. This protocol enables the data owner to categorize their material privately without having direct access to the learned model.

## 2.6.5. RQ5: How can artificial intelligence be used to predict medical pathologies?

With the aim of understanding and treating disease by integrating multimodal or multi-omics data from an individual to make patient-appropriate decisions, precision medicine is an emerging approach to clinical research and patient treatment. To categorize and predict outcomes for both people and communities, multi-omics data analysis and processing techniques, including novel deep learning models, are crucial [39].

People today are affected by a variety of diseases, making early-stage disease prediction a crucial task. Unfortunately, a doctor cannot accurately anticipate a disease based just on its symptoms, making accurate disease prediction the most challenging effort. To solve this issue, data mining is crucial in disease prediction. We point out that the study by Dahiwade et al [40] discovered that the CNN algorithm is more precise and efficient in predicting diseases than KNN. In a different study, Shipe et al [41] reviewed the steps involved in creating a logistic regression risk prediction model, including everything from selecting a data source and predictor variables to analyzing the model's performance and enhancing clinical decision-making. Finally, Nusinovici et al [42] demonstrated that logistic regression is equally effective as ML models at predicting the risk of major chronic diseases with low incidence and simple clinical predictors.

Various experiments in the medical sector have been applied to various fields of health. RAMZAN et al [43] discovered that using functional magnetic resonance imaging and advanced deep learning methods to classify and predict neurodegenerative brain disorders such as Alzheimer's disease is beneficial for clinical decision- making and has the potential to aid in AD prognosis.

Clinical notes and UMLS resources are effective and crucial tools for estimating mortality in diabetic critical care patients. In a recent study, Ye et al [44] created several predictive models to interpret the mortality of diabetic patients. The goal of the study was to estimate mortality risk using the UMLS resources, which included machine learning and NLP methods.

We thus note the intervention of ALAWAD et al [45] who approved that the use of UMLS vocabulary resources to enrich the word inclusions of CNN models have systematically surpassed the Convolutional Neural Network (CNN) models without integration of pretrained words to detect cancerous pathology. Furthermore, WANG et al [46] used a deep CNN to construct an automated tumor region detection system for lung cancer pathology images, which allowed them to gain fresh insights into the association between tumor morphology and patient prognosis.

Due to the rapid expansion of digital technology, CHO et al [47] provided an effective model for future research by developing a mood prediction algorithm using machine learning, which may even enhance the prognosis of patients with mood disorders by allowing for a practical clinical application. In addition, through a review of existing research in the literature referring to Data Mining techniques and algorithms in mental health, ALONSO et al [48] found that the application of Data Mining techniques to disorders like depression can be extremely beneficial in terms of clinical decision-making, diagnosis prediction, and patient quality of life.

The majority of recent work on the prediction of clinical pathologies has been on the prediction of diabetes. LJUBIC et al [49] proved that the RNN model was the best choice for the electronic medical record data type, to predict diabetes complications. The last intervention in this context is that of RASMY et al [50] who evaluated the logistic regression and a recurrent neural network as well the UMLS for the prediction of the risk of heart failure in diabetic patients.

# 3 Proposed approach

After analyzing and reviewing previously published papers that answer the various research questions of our systematic review by exploiting the most recent technologies such as big data, artificial intelligence, machine learning and deep learning for the improvement of health care, we noted that the key to an accurate prediction of medical pathologies in a better time is a well-defined reliable process that covers all the steps that precede the prediction starting with the collection of medical data, then the pre-processing of the data collected, followed by the mapping of the medical data from the data collected previously pre-processed, then the classification and the clustering of the patient profiles in order to be able to predict the medical pathologies within a significant period of time with high precision. To do this we were able to propose a general approach to predict clinical pathologies with precision in a better time, our approach is based on five main steps:

## 3.1 Collecting Data:

The first step (Figure 2) of our proposed approach is based on the combination of medical data (electronic medical records, doctor's observations...) in different formats (PDF, RTF, HTML, XML...), a "Collection" function that takes into account input the path of the file directory (DME) is set up to return us the type of each file and its path, subsequently a "DATA" class is created containing the path to each file and its format.

To reduce the complexity of massive medical data, a "Data Analyzer" function has been implemented, takes the "DATA" class as input, and divides all the files along two main axes:

- **Text (RTF, PDF):** Contains all unstructured files that require further analysis and processing.

- **Relational DB (.csv, .db):** Contains all semi-structured files that require less analysis and processing.

A set of wrappers will be set up to execute the sub-queries in the different data sources and transform the results of the sub-queries into JSON documents which will be stored subsequently in the "Data Set".

At the end of this data collection step, a "Parser" will be set up, it takes the data from the Dataset as input and returns it as output:

- **Metadata:** This contains the most important information about each file to make it easier to find and archive.

- **The new "EMR" Dataset:** Contains all the fields necessary for the processing and analysis (medical data) of all patients to facilitate data mining, reduce complexity and time by just analyzing the data needed.

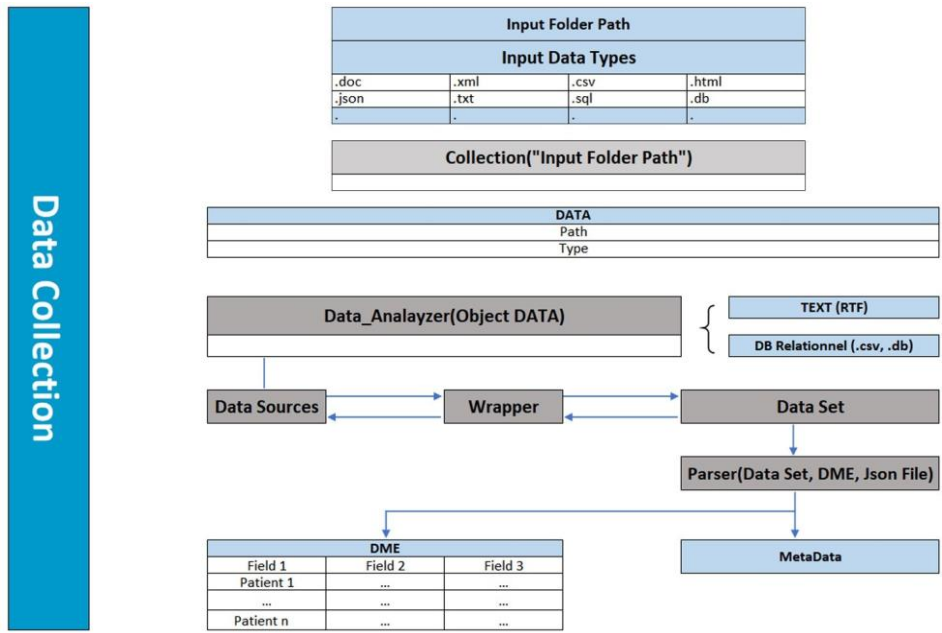

Figure 2: Collecting Data

## 3.2 Preprocessing Data:

The second step (Figure 3) of our proposed approach concerns the application of the preprocessing process on the "DME" Dataset. This process consists of data cleaning, integration, reduction, and transformation of medical data using the python-written Natural Language Toolkit (NLTK) [51] software library, which is distributed under the General Public License (GPL) as a collection of program modules, datasets, and tutorials to support the study and teaching of computational linguistics and natural language processing, and which supports several features like transparent syntax, effective string management features, and simplicity, using the treebank word tokenizer and POS tagger. The NLTK library used allows to automatically process with better performance to have in output a structured and preprocessed "SDME" Dataset.

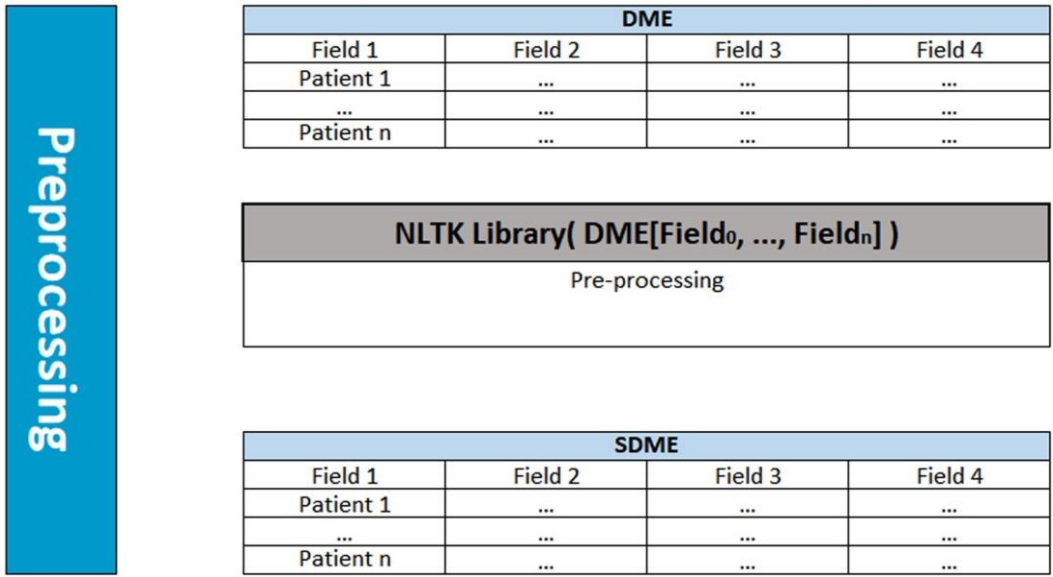

Figure 3: Preprocessing Data

### 3.3 Mapping Medical Data:

The third step of our proposed approach (Figure 4) is to map the biomedical terms contained in the structured data set "SDME" Using the "UMLS" API, which combines many health and medical vocabularies and standards into a set of files and software to facilitate interoperability with computer systems, as well as the biomedical entity "GATE (Bio-YODIE)" API which is a named entity recognition and disambiguation system that identifies different types of biomedical entities in the text and tries to associate them with the most appropriate conceptual tag of concept in UMLS to discover the concepts of the meta thesaurus mentioned in the text, allowing a more sophisticated understanding to have as an output a new Dataset "Mapped SDME" which contains the fields:

- **Mandatory terms:** Represents a list of words or symptoms, each of which is mapped to the UMLS metathesaurus.

- **Concept Extraction:** A concept is the meaning of medical terms and each meaning contains different names.

- **Semantic Type Extraction:** Another source of UMLS knowledge is the semantic network, which contains 135 Semantic Types and 54 semantic relations to classify and categorize all the concepts represented in the UMLS metathesaurus.

- **Entity type extraction:** An entity type displays the parent relationship of the concept; the meaning of the concept is presented in a more standard and obvious way compared to the semantic type.

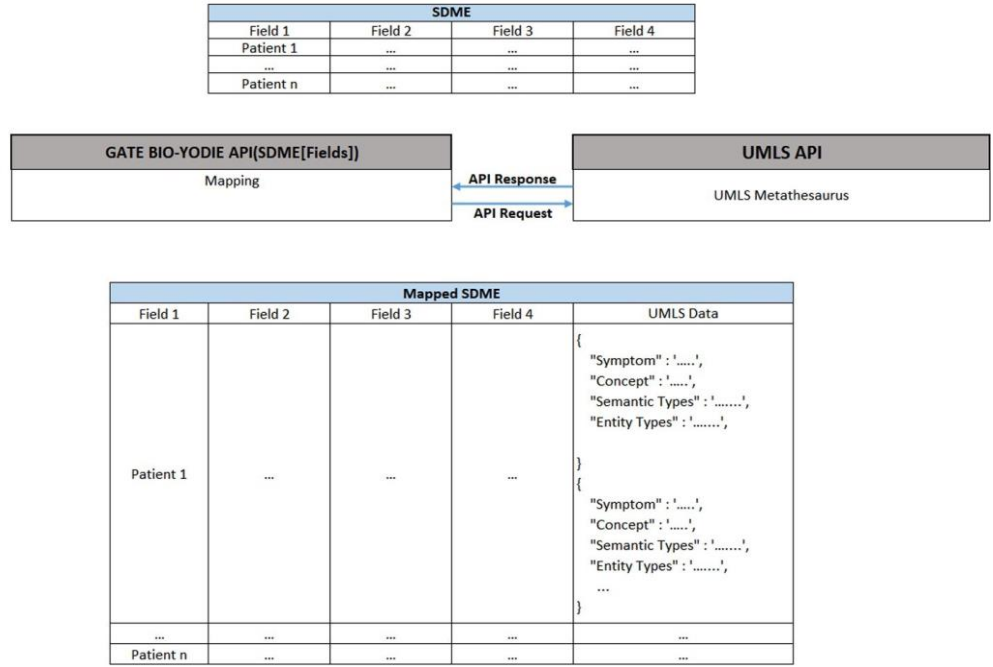

Figure 4: Mapping Medical Data

## 3.4 Classification and Clustering:

The fourth step (Figure 5) of our proposed approach concerns the classification and clustering of patient profiles, so a "Clustering" function that takes as input the data stored in the "Mapped SDME" Dataset is set up to define the different clusters of profiles/patients based on the rules laid down beforehand by the healthcare professionals using the k-means clustering algorithm typical in data mining which is commonly used in data mining to divide a big amount of data into k different clusters , this algorithm categorizes objects based on whether or not they match the requirements of a particular class [52]. In our approach the K-means algorithm used to assign each patient to one or more clusters according to the data in the "UMLS Data" field to have as output a new "Clustered Mapped SDME" Dataset which contains the "Cluster" field made up of the different cluster (s) of each patient.

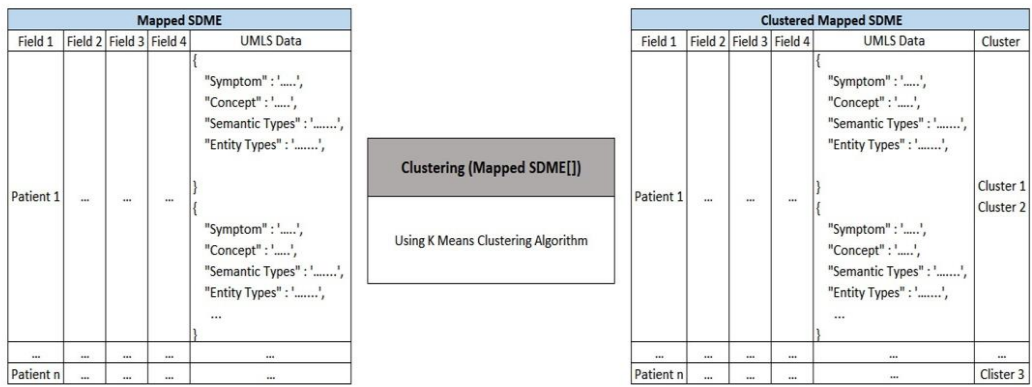

Figure 5: Classification and Clustering

### 3.5 Pathology Prediction:

The last step (Figure 6) of our proposed approach concerns the development of an Artificial Intelligence model using the famous Recurrent Neural Networks (RNN) algorithm, to analyze data streams using hidden units that focus on the modeling and processing of sequential data; which consists of loops and memory so that they remember old calculations, unlike any other feedforward neural network [53], the proposed Artificial Intelligence model takes as input the data taken from the dataset "Clustered Mapped SDME" and produces the final "Pathology Prediction" dataset which contains the following three new fields:

- **"Pathologies" field:** allows the prediction of the most frequent and serious pathologies.

- **"Best Prediction" field:** allows pathologies to be predicted more precisely.

- **"Best Precision" field:** allows pathologies to be predicted within a significant timeframe.

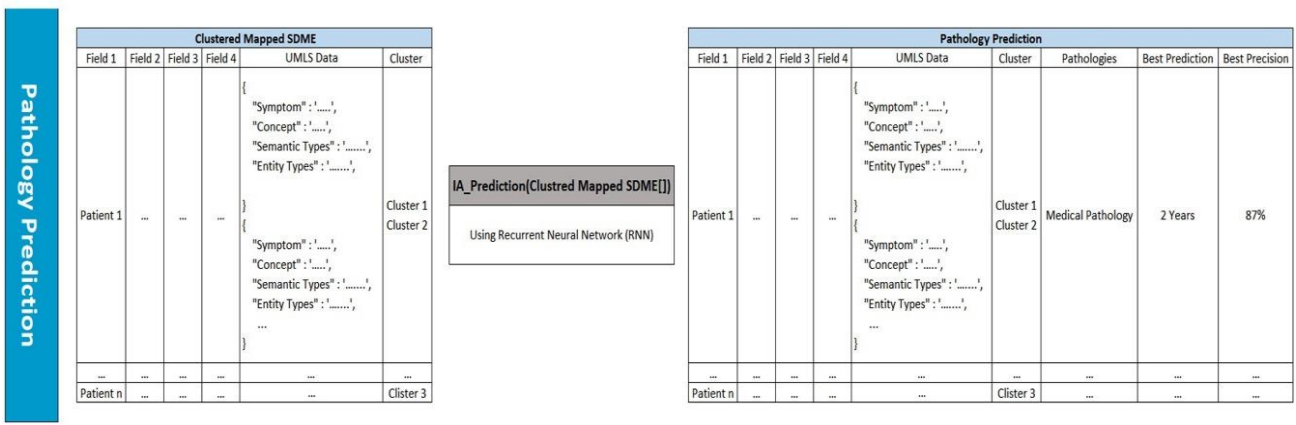

Figure 6: Pathology Prediction

## 4 Discussion and limitations

Our article's major goal was to examine the best examples of research that demonstrates the value of utilizing the most recent technologies including big data, artificial intelligence, machine learning, and deep learning, to enhance healthcare by predicting medical pathologies. According to our survey, many researchers are currently interested in taking advantage of the digital revolution and applying computer science-derived methods to the medical industry in order to raise the standard of medical care on the one hand, and suggesting innovative medical treatments that would ease the lives of patients and medical staff by accurately and quickly predicting the most prevalent medical diseases on the other. To achieve this, our article suggests a general approach that can be used to predict various medical pathologies, which is more than the systematic review that is presented. The main benefit of our proposed approach is that it suggests a complete process that is well detailed and covers all the stages of the data collection until the prediction of the medical pathologies in a notable delay.

Based on this discussion, it can be concluded that it is important to note that several published researches are unlikely to have been included in our systematic review due to the quickly evolving literature in the field of medical informatics. The elaborate study analyzed studies from three databases, including Scopus, PubMed, and Google Scholar. Future studies could compare the findings using papers from various databases like Medline and Web of Science.

# 5 Conclusion

E-health is developing exponentially and digital technology is at the heart of innovation, both in clinical research and in-patient care and support. The use of big data in the medical sector offers a multitude of new possibilities and changes the way we collect data, in particular because health data is far too large and complex to be exploited by traditional methods of management and treatment appointment.

Our article develops new evidence that the key to an accurate prediction of medical pathologies in a better time is a well-defined reliable process that covers all the steps that precede the prediction by employing the most recent recommended technologies such as big data, intelligence, machine learning and deep learning. Forty-nine selected articles provided us with good arguments to represent the ideas answering five research questions. These results will help to develop researchers' understanding of data collection or preprocessing methods in health informatics, data mapping or clustering techniques in health informatics, and how artificial intelligence can be used to predict health informatics medical pathologies.

Based on the systematic review elaborated in this article, we succeeded in proposing a general approach applicable to all fields of medicine, which is based first of all on the collection of unstructured and semi-structured data from different hospitals. and cabinets, then the preprocessing of these data carried out by the application of data mining techniques on the medical data collected using the python library Natural Language Toolkit (NLTK), thereafter the medical data is mapped by referring to the Unified Medical Language System (UMLS) to facilitate data analysis and reduce complexity and time consumed, then we apply clustering to categorize the different pathologies and facilitate patient profiling and finally an artificial intelligence is developed to predict the most frequent pathologies with better precision in a noticeable time.

## Declarations

- **Funding:** No funding was received to assist with the preparation of this manuscript.
- **Conflicts of interest:** The authors declare that they have no conflicts of interest.
- **Ethical Approval:** This article does not contain any studies with human participants or animals performed by any of the authors.

## Author Information

**Biography:**

- Chaimae Taoussi is a PhD Student at Sultan Moulay Slimane University. Her latest published paper is "Predicting Psychological Pathologies from Electronic Medical Records".

- Imad Hafidi is a research professor at Sultan Moulay Slimane University. His latest published paper is "A novel secure data aggregation scheme based on semihomomorphic encryption in wsns".

- Abdelmoutalib Metrane is a research professor at Sultan Moulay Slimane University. His latest published paper is "Robust approach for blind separation of noisy mixtures of independent and dependent sources".

**Affiliations:**

- Chaimae Taoussi, Laboratory of Process Engineering Computer Science and Mathematics, University Sultan Moulay Slimane, Beni Mellal, Morocco.

- Imad Hafidi, Laboratory of Process Engineering Computer Science and Mathematics, University Sultan Moulay Slimane, Beni Mellal, Morocco.

- Abdelmoutalib Metrane, Laboratory of Process Engineering Computer Science and Mathematics, University Sultan Moulay Slimane, Beni Mellal, Morocco.

**Contributions:**

- Authors Chaimae Taoussi, Imad Hafidi, and Abdelmoutalib Metrane contributed equally to this work. All authors have read and approved the final manuscript.

**Corresponding author:**

- Correspondence to Chaimae Taoussi.